\def\year{2018}\relax
\documentclass[letterpaper]{article} 
\usepackage{aaai18}  
\usepackage{times}  
\usepackage{helvet}  
\usepackage{courier}  
\usepackage{url}  
\usepackage{graphicx}  
\usepackage{epsfig}
\usepackage{graphicx}
\usepackage{amsmath}
\usepackage{amssymb}
\usepackage{subfigure}
\usepackage{float}
\DeclareMathOperator*{\argmin}{arg\,min}
\newcommand{\eref}[1]
{(\ref{#1})}

\newcommand{\fref}[1]
{\mbox{Fig.~\ref{#1}}}

\begin{document}
%
\title{Self-Reinforced Cascaded Regression for Face Alignment}
\author{Xin Fan$^{1,2,}$\thanks{Corresponding Author.}, \ Risheng Liu$^{1,2}$, \ Kang Huyan$^{1,2}$, \ Yuyao Feng$^{1,2}$, \ Zhongxuan Luo$^{1,2,3}$\\
$^1$DUT-RU International School of Information Science \& Engineering, Dalian University of Technology, Dalian, China\\
$^2$Key Laboratory for Ubiquitous Network and Service Software of Liaoning Province, Dalian, China\\
$^3$School of Mathematical Science, Dalian University of Technology, Dalian, China\\
\{xin.fan, rsliu, zxluo\}@dlut.edu.cn, \  huyankang@hotmail.com \ yyaofeng@gmail.com
}

\maketitle
\begin{abstract}
Cascaded regression is prevailing in face alignment thanks to its accuracy and robustness, but typically demands manually annotated examples having low discrepancy between shape-indexed features and shape updates. In this paper, we propose a self-reinforced strategy that iteratively expands the quantity and improves the quality of training examples, thus upgrading the performance of cascaded regression itself. The reinforced term evaluates the example quality upon the consistence on both local appearance and global geometry of human faces, and constitutes the example evolution by the philosophy of ``survival of the fittest". We train a set of discriminative classifiers, each associated with one landmark label, to prune those examples with inconsistent local appearance, and further validate the geometric relationship among groups of labeled landmarks against the common global geometry derived from a projective invariant. We embed this generic strategy into typical cascaded regressions, and the alignment results on several benchmark data sets demonstrate its effectiveness to predict good examples starting from a small subset.
\end{abstract}


\section{Introduction}
 Face alignment, aiming at accurately and robustly localizing facial landmarks, plays a key role to many automatic facial analysis tasks including face recognition, expression recognition, attribute analysis, and animation. Recently, cascaded regression has become one of the most popular approaches to face alignment due to its accuracy and robustness~\cite{ren2014face,xiong2013supervised,kowalski2017deep}. This approach learns a series of regressors between shape-indexed features and shape updates or gradients from a set of manually labeled face images. Inevitably, the performance of cascaded regression highly depends on the quantity and quantity of training examples. The quantity of unlabeled facial images is not a problem in this 'big-data' era, but example labeling and the quality of labels are still critical. In this study, we focus on these critical issues for cascaded regression.

Despite of its great success, the discrepancy or mismatch between limited training examples and the huge solution space typically downgrades the stability and accuracy of cascaded regression. One typical treatment is to divide the original shape space into smaller sub-spaces~\cite{zhuunconstrained,Tuzel2016}. Researchers also attempt to group relevant input features for mitigating mismatches~\cite{cao2014face,ren2014face}. The cascade Gaussian process (GP) regression trees find input features showing consistent appearance through GP kernel functions~\cite{lee2015face}. The common strategy of these methods lies in that they `tighten' the correlation between input feature and target shape from the perspective of \emph{local appearance}.

Alternatively, researchers resort to the \emph{global geometry} (shape) among facial landmarks in order to address the discrepancy issue. Martinez \emph{et al.} embed nonparametric Markov networks~\cite{martinez2013local}, while Liu \emph{et al.} incorporate sparse shape constraints into regression~\cite{liu2016dual}. In addition to these explicit shape models, Li \emph{et al.} discover the common geometry shared by human faces using a projective invariant, called characteristic number (CN), and append this geometric regression to appearance~\cite{Li2015}. These various forms of facial geometric representation are able to regularize the regression, and thus improve the robustness of alignment.


It is commonly accepted in the machine learning (ML) community that training examples are central to any ML algorithms including regression. Unfortunately, the aforementioned alignment algorithms pay more attention to the regression mechanism, instead of data itself, to tackle the issue arisen from data discrepancy. Targeting at data preparation for training and validating regressors, Sagonas \emph{et al.} develop a semi-automatic tool to annotate facial landmarks~\cite{Sagonas_2013_300w}, but how these annotations may affect regression is untouched in their study. Antonakos \emph{et al.} generate bounding boxes as face labels and validate these labels in the context of linear parametric models but not more complex cascade regression~\cite{antonakos2014automatic}. Recently, Zhang \emph{et al.} develop a complicated deep network to leverage face annotations across data sets~\cite{zhang2015leveraging}. Nevertheless, a general framework is still highly demanded to fuse the discovering and upgrading training examples of low discrepancy into cascaded regression for face alignment.

Self-reinforcement refers to ``a process whereby individuals control their own behavior by rewarding themselves when a certain standard of performance has been attained or surpassed"~\cite{Artino2011}. In this paper, we propose self-reinforced cascaded regression that upgrades itself through minimizing an objective function analogous to meeting the performance standard. The optimization process iteratively updates \emph{example labeling}, \emph{sample survival}, and \emph{regression} in one framework as shown in~\fref{fig:FirstFig}. The process starts from predicting unlabeled faces by the regression trained from a small number of labeled examples, and then evaluates the consistence of predicted labels on both \emph{local appearance} and \emph{global geometry} of human faces.
Those survived examples are fed to train an upgraded regression. This process iteratively runs until convergence, yielding the cascaded regression for accurate and robust alignment.

The objective in our framework is not directly defined on the consistence between predicted labels and the ground truth as typical semi-supervised learning~\cite{zhu2009introduction} that has the risk of overfitting, but is derived from indirect consistency with local appearance and global geometry. This independence on regressors is so general to generate the self-reinforced versions of various cascaded regression algorithms. We demonstrate that our strategy is able to automatically predict and find good examples starting from a subset as small as \emph{one hundred} for typical regressors~\cite{ren2014face} and~\cite{zhu2015face}, and even deep networks~\cite{kowalski2017deep}. These self-reinforced regressions output comparable accuracy with the state-of-the-art on the 300W set consisting of the test sets of LFPW and Helen~\cite{Le2012} when only a small fraction of labeled examples are available, validating its effectiveness.
\section{Related Work}

In this section, we review recent advances on labeling or generate examples in the machine learning community.

Semi-supervised learning attempts to use unlabeled data for performance improvements of classifiers trained by a small number of labeled examples~\cite{zhu2009introduction}. It has made great progress on solving the discrete classification problems in this decade~\cite{li2013low,li2015towards}. However, it is nontrivial to directly bring the semi-supervised algorithms for discrete problems to cascaded regression where target shape updates are continuous and the solution space is quite huge. Self-paced learning (SPL), falling in the category of semi-supervised learning, include training samples in an easy-to-complex fashion~\cite{jiang2014easy,singh2015selecting}. Our approach shares commons with SPL on example selection embedded in the training process, differing in that our objective is general and decoupled from the training objective.

Generative adversarial network (GAN) ~\cite{goodfellow2014generative} is able to generate visually realistic images by competing two deep networks, a generator and a discriminator. Recently, GAN finds wide applications in many low level image processing tasks such as super-resolution~\cite{ledig2016photo} and image attribute transfer~\cite{huang2017beyond}. Semi-supervised learning can also be combined with GAN in order to improve the realism of a simulator's output while preserving the annotation information~\cite{shrivastava2016learning}
. Our example prediction and survival share the similar spirit with the generative and discriminative processes in GAN, respectively. But GAN has to initialize from a relatively larger number of examples to train two deep networks as the generator and discriminator, and provides no explicit regressor as self-reinforced regression does.


\section{Self-Reinforced Cascaded Regression}
\begin{figure}[H]
\begin{center}
   \includegraphics[width=0.9\linewidth]{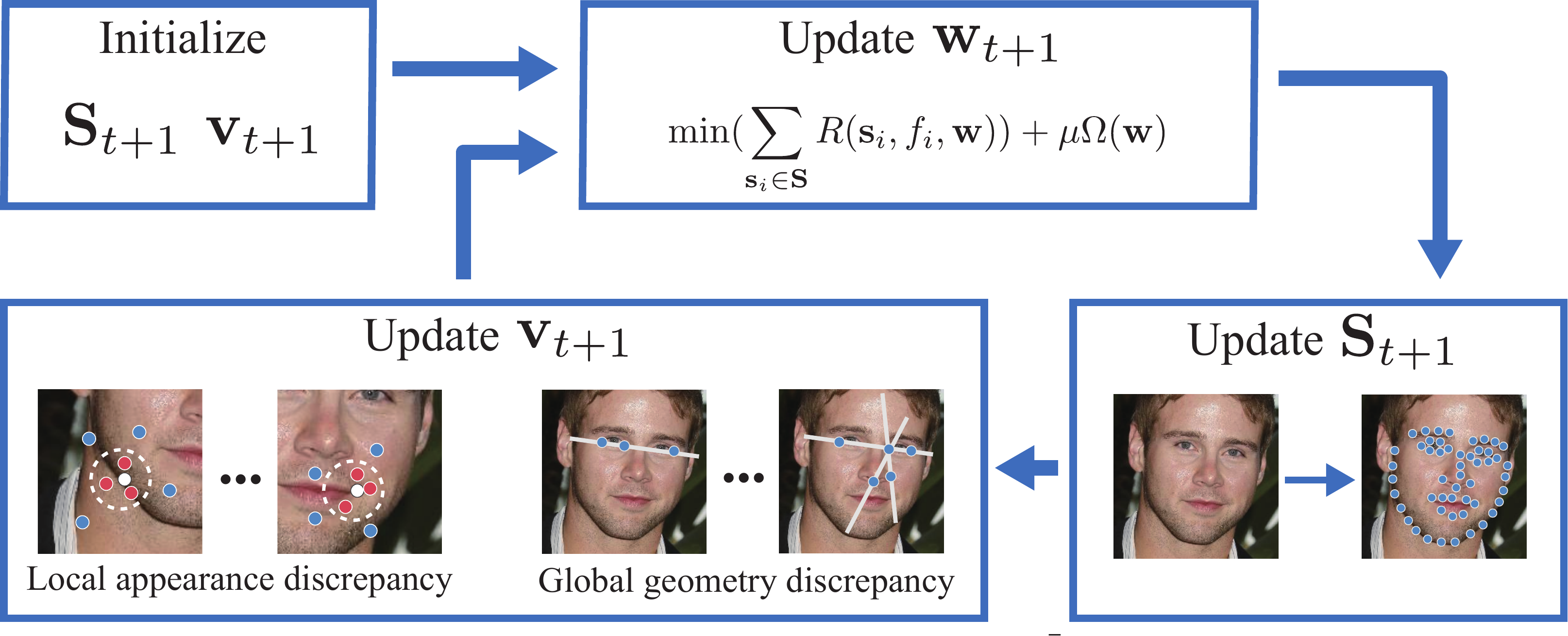}
   \end{center}
   \caption{Overview of our self-reinforced cascaded regression, forming a closed loop with label prediction $\mathbf{S}$ and survival $\mathbf{v}$ as well as regression upgrading $\mathbf{W}$.}
  \label{fig:FirstFig}
\end{figure}

We describe our self-reinforced cascaded regression that defines an objective function with a local appearance and a global geometry discrepancy to iteratively expand the training set and simultaneously upgrade the regressor as shown in~\fref{fig:FirstFig}
\subsection{General formulation}
We attempt to devise a general formulation where the self-reinforcement is embedded with cascaded regression. Typical cascaded regression minimizes a loss function $R(\mathbf{s}_i, f_i, \mathbf{w})$, where $\mathbf{s}_i$ is the
annotated shape of the $i$th sample in the training set $\mathbf{S}$. The symbol $f_i$ indicates the shape-indexed feature of the $i$th sample image, and $\mathbf{w}$ denotes the parameters of the learnt regressor. We denote $\Omega(\cdot)$ as the regularization term and $\mu$ as a hyper parameter, and thus have a general representation for cascaded regression as follow:

\begin{equation}
\min(\sum_{\mathbf{s}_i \in \mathbf{S}}R(\mathbf{s}_i, f_i, \mathbf{w})) + \mu\Omega(\mathbf{w}).
\label{eq:face_align}
\end{equation}

Given the cascaded regression representation~\eqref{eq:face_align},
we impose a regularize term to formulate the iterative reinforcement of predicted examples as:
\begin{equation}
\begin{aligned}
&(\mathbf{w}_{t+1}, \mathbf{v}_{t+1}, \mathbf{S}_{t+1}) = \\
&\argmin_{v_i\in\{0,1\}^n}(\sum_{\mathbf{s}_i \in \mathbf{S}_t}v_iR(\mathbf{s}_i, f_i, \mathbf{w}_t) + \mu\Omega(\mathbf{w}_t)
-\frac{1}{\alpha}\sum^n_{i=1}v_i),
\label{eq:objective_func}
\end{aligned}
\end{equation}
where the subscript $t$ indicates the $t$th iteration. The training set $\mathbf{S}_t$ for the regression $R$ includes either manually labeled or originally unlabeled examples with predicted annotations. The vector $\mathbf{v}$ consists of the binary $v_i$ that indicates whether the $i$th sample is accurately labeled or not, and the parameter $\alpha$ is a weight that determines the number of survived samples. The increase of $\alpha$ during the iteration leads to including more samples for regression.

The objective function~\eref{eq:objective_func} embraces the regression $\mathbf{w}$, shape labels $\mathbf{s}$ and example selection $v$ into one general framework whose optimization brings the joint upgrading of all these factors. Consequently, the optimization of this objective forms a complicated problem with the mixture of continuous and discrete variables. We resort to an iterative approximation to find the solution of~\eref{eq:objective_func}. First, we fix $\mathbf{v}_{t}$ and $\mathbf{S}_{t}$ to find the optimal regression parameters $\mathbf{w}_{t+1}$. The problem~\eqref{eq:objective_func} degrades to conventional cascaded regression~\eqref{eq:face_align}, e.g.,~\cite{ren2014face} and~\cite{zhu2015face} as detailed in the next section. For initialization, $\mathbf{v}_{t}$ is set to $1$ if the sample is manually labeled otherwise $0$.

Once the trained regression $\mathbf{w}_{t+1}$ is available, we are able to predict the unlabeled or to update the labeled subset. Given fixed $\mathbf{v}_t$ and $\mathbf{w}_{t+1}$ in~\eqref{eq:objective_func}, the updating of example labels $\mathbf{S}_{t}$ becomes:
\begin{equation}
\min(\sum_{\mathbf{s}_i \in \mathbf{S}}R(\mathbf{s}_i, f_i, \mathbf{w})).
\label{eq:face_align_2}
\end{equation}
This minimization is equivalent to perform a prediction by applying the learned cascaded regression. This updating is so important in our self-reinforced regression that the process does not only expand the example quantity but also improves the labeling accuracy by the regression trained from the survived examples in the previous iteration.

Finally, we update $\mathbf{v}_{t}$ with  $\mathbf{S}_{t+1}$ and $\mathbf{w}_{t+1}$ fixed by degenerating~\eqref{eq:objective_func} to:
\begin{equation}
\min(\frac{1}{\alpha}\sum^n_{i=1}v_i).
\label{eq:face_align_2}
\end{equation}
We compute the indicator $\mathbf{v}_{t+1}$ upon local appearance and global geometry of human faces:
\begin{equation}
v_i =
\left\{
\begin{aligned}
1&& -(\log a_i + \lambda\log g_i) <  \alpha\\
0&& \text{otherwise}
\end{aligned},
\right.
\end{equation}
where the parameter $\lambda$ weighs appearance and geometry. The value $a_i$ is derived from local appearance indicating how accurately an individual landmark labels, and $g_i$ indicates how a group of predicted labels satisfies the common geometry of human faces. The calculation of this new regularization term is independent to the regression $R$, providing the generalization for various regression algorithms.

\emph{Remark:} The calculation of $a_i$ and $g_i$ acts as the goodness evaluation of individuals (examples), and hence initiates adjusting the behavior (accuracy) of individuals and that of cascaded regression for the next iteration, constructing the self-reinforcement process. The binary indicator $\mathbf{v}_i$ specifies whether one label survives or not, implying the well-known law of nature ``survivor of the fittest". As nature evolves repeatedly, our self-reinforced cascaded regression iteratively upgrades from a small subset of labels until $a_i$ and $g_i$ are stable as shown in~\fref{fig:FirstFig}.

\subsection{Local appearance discrepancy}

\begin{figure}[!htb]

\subfigure[]{
\includegraphics[width=.3\linewidth]{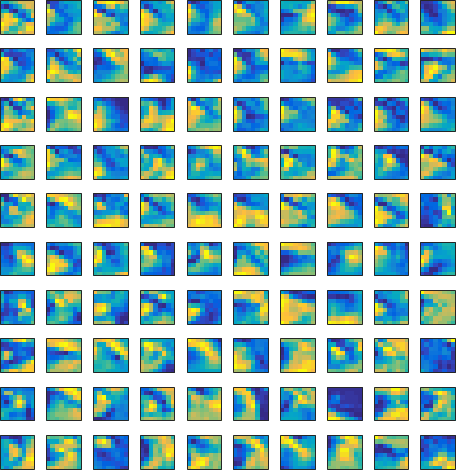}
}
\subfigure[]{
\includegraphics[width=.3\linewidth]{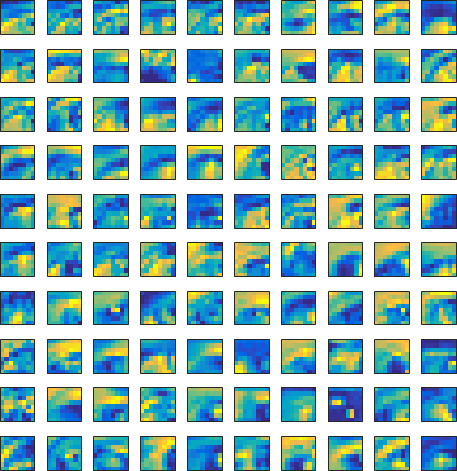}
}
\subfigure[]{
\includegraphics[width=.3\linewidth]{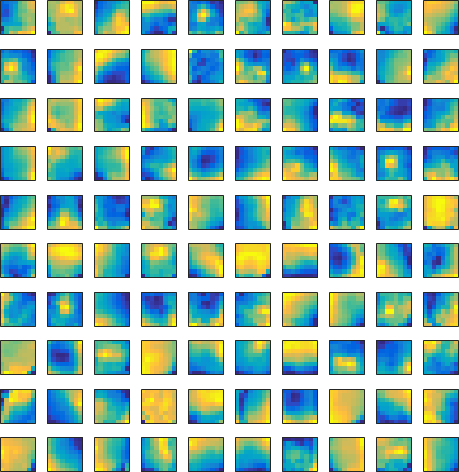}
}
\caption{Local patches around (a) right corner the right eye, (b) upper boundary of the right eye, and (c) nose tip.}
\label{fig:local_patches}
\end{figure}

We define $a_i$ as the discrepancy (similarity) among the shape-indexed features (concatenating HOG~\cite{dalal2005histograms} and FREAK~\cite{alahi2012freak}) associated with an individual landmark. Figure~\ref{fig:local_patches} demonstrates the patches around three landmarks, i.e., the right corner, the upper boundary of the right eye, and the nose tip, from manually labeled images. The patches around the same landmark exhibit similar appearance, while greatly different from the other landmarks. Hence, the consistency of local patches around a landmark is able to indicate the accuracy of the labeled position.

\begin{figure}[!htb]
\centering
\includegraphics[width=0.8\linewidth]{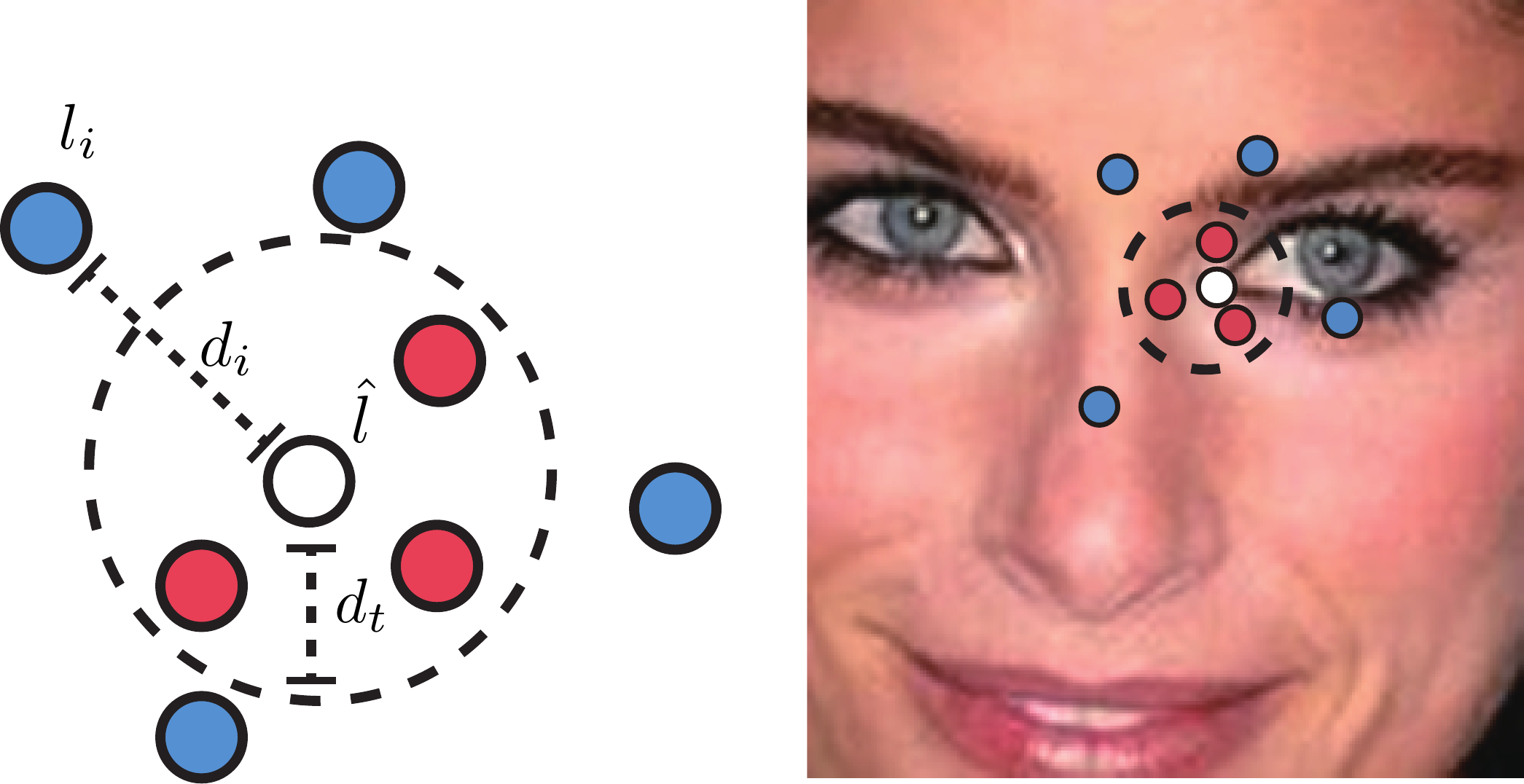}
\label{fig:gen_landmark_ex}
\caption{Generate training samples for the local appearance classifier.}
\label{fig:gen_landmark}
\end{figure}

We take a straightforward technique to train an offline naive Bayes classifier that discriminates those labels with inconsistent neighboring appearance. We generate the positive and negative samples for training the classifier from the originally labeled subset by assuming that labeled and predicted landmarks are normally distributed. Hence, we randomly perturb the ground truth labels with a normal distribution, and compute the distance $d_i$ between the ground truth $\hat{l}$ and the perturbed landmark $l_i$. The feature around the landmark whose $d_i$ is less than a threshold $d_t$ (related to the standard deviation of the Gaussian distribution) is taken as one positive sample for the classifier, others as the negative. This generation scheme is illustrated in~\fref{fig:gen_landmark}, where the white dot denotes the ground truth, the red ones stand for positive samples and the blue for negative ones.

Given a predicted landmark, we apply the trained classifier to determine whether the landmark is a valid prediction, and evaluate the local appearance discrepancy $a_i$ for a predicted (or labeled) example as the portion of valid landmarks in the example:
\begin{equation}
a_i = \frac{\sum_{\mathbf{p}\in L_i}\arg\!\max_{c\in \mathcal{Y}} P(c) \prod_{k=1}^m P(p_k|c)}{|L_i|}
\end{equation}
The symbol $L_i$ denotes the set of local features for all landmarks in the $i$th sample, $|L_i|$ is the number of landmarks,and the local feature vector $\mathbf{p}$ has $m$ components. The classifier output $\mathcal{Y}$ is binary, where $c=1$ indicates a valid landmark and zero stands for an invalid one.

\subsection{Global geometry discrepancy}

The above discrepancy can only reflect the local feature consistency around a landmark. We use the intrinsic facial geometry given by a projective invariant, named the characteristic number (CN)~\cite{fan2015fiducial}, to evaluate the discrepancy of predicted or labeled examples.

\begin{figure}[!htb]
\centering

\includegraphics[width=.9\linewidth]{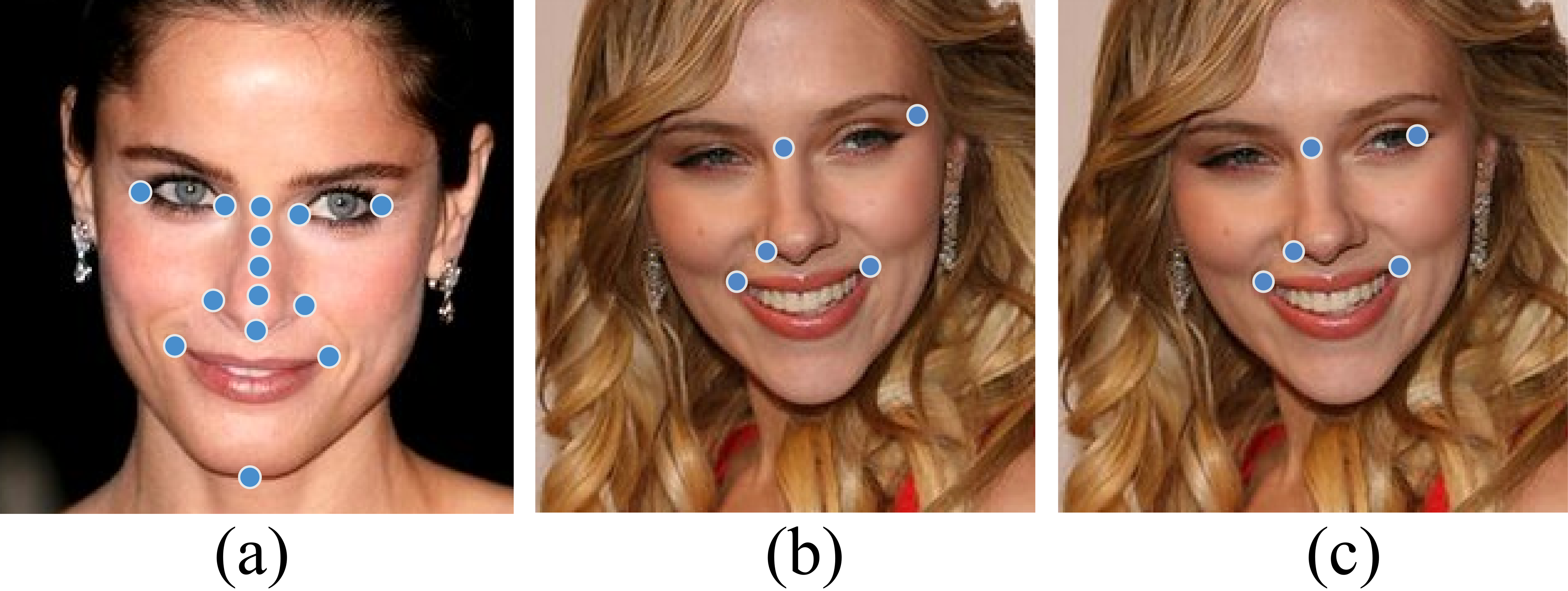}

\caption{CN values reflect landmark geometry: (a) all chosen landmarks to generate point combinations for CN calculation, (b) one combination with the groundtruth, $CN=0.0204$, and (c) one combination with an inaccurately labeled landmark, $CN=0.0479$.}
\label{fig:CN_landmarks}
\end{figure}



Fan~\emph{et al.} discover the common geometry on 8 landmarks~\cite{Li2015}. Herein, we are considering to label and select examples with 68 landmarks. Unfortunately, it is prohibited for us to investigate all combinations of these 68 landmarks. We pick 14 landmarks that are all stably presented in all face examples, shown as the blue points in~\fref{fig:CN_landmarks}(a). We enumerate all possible three-point, five-point and six-point\footnote{Four points cannot construct a projective invariant.}  combinations of these 14 landmarks, and then calculate the CN values of these combinations on all available samples. If a combination presents one common CN value with low standard deviation for all sample images, we set the value as the \emph{intrinsic value} reflecting the common geometry underlying this landmark combination. Figure~\ref{fig:CN_landmarks}(b) and (c) show one sample with correctly labeled landmarks and another with an inaccurately labeled landmark, respectively. Their CN values are quite different. We have to emphasize that this process seeking combinations with stable intrinsic values only runs once for a large face data set. We verify the CN values of predicted landmark annotations on these fixed combinations in the iterative selecting process.


%
%

It is reasonable to regard a set of landmark annotations (labels) as valid when its CN value falls within a range around its corresponding intrinsic value, recorded as [$c^{(min)}_k$~$c^{(max)}_k$]. Accordingly, the discrepancy $g_i$ for the global geometry is given below:

\begin{equation}
\begin{aligned}
&g_i = \frac {\sum_{c_k\in C} p_{c_k}} {|C|}\\
&\text{where ~} p_{c_k} =
\left\{
\begin{aligned}
1 & & c^{(min)}_k<= c_k <=c^{(max)}_k\\
0 & & \text{otherwise}
\end{aligned},
\right.
\end{aligned}
\end{equation}
$c_k$ is the $k$th combination of CN values in the $i$th sample, and $|C|$ is the total number of combinations, each of which can give one intrinsic value.

\section{Alignment Algorithms}

The last regular term in~\eref{eq:objective_func} is independent on the choice of regression, and thus it is ready to embed the proposed algorithm into any cascaded regression algorithms. In this section, we exemplify the embedding to two algorithms LBF~\cite{ren2014face} and CFSS~\cite{zhu2015face} that balance accuracy and efficiency.

In every iteration, LBF have two updating stages: one for learning local binary features $\Phi=[\phi_1, \phi_2, \dots,\phi_n]$, and the other for global linear regression $\mathbf{W}^t$. We pose the learning for the first stage as the minimization of the objective function~\eqref{eq:local_LBF_eq}, where $\pi_l \circ \Delta \mathbf{z}_i$ is the ground truth 2-dimensional offset of the $l$th landmark in the $i$th  training sample. $I_i$ is the facial image corresponding to $i$ sample:

\begin{equation}
\begin{aligned}
\min_{w_l, \phi_l}
\sum_{
\mathbf{s}_i,\Delta\mathbf{s}_i
}
||\pi_l \circ \Delta \mathbf{s}_i - w_l \phi_l(\mathbf{s}_i, I_i)||_2^2, \\
\text{where}~\mathbf{s}_i \in \mathbf{S}_{t-1},\Delta\mathbf{s}_i \in \mathbf{S}_t.
\end{aligned}
\label{eq:local_LBF_eq}
\end{equation}
Subsequently, we transform~\eqref{eq:objective_func} into~\eqref{eq:objective_func_LBF} in order to obtain the linear regression $\mathbf{W}^t$ in LBF and combine it into our formulation.
\begin{equation}
\begin{aligned}
&(\mathbf{w}_{t+1}, \mathbf{v}_{t+1}, \mathbf{S}_{t+1}) = \\
&\argmin_{v_i\in\{0,1\}^n}(\sum_{
\mathbf{s}_i,\Delta\mathbf{z}_i
}v_i||\Delta \mathbf{s}_i - \mathbf{w}_t \Phi(\mathbf{s}_i, I_i)||_2^2 \\
&+ \mu ||\mathbf{w}_t||_2^2,
-\frac{1}{\alpha}\sum^n_{i=1}v_i), \\
&\text{where}~\mathbf{s}_i \in \mathbf{S}_{t-1},\Delta\mathbf{s}_i \in \mathbf{S}_t
\end{aligned}
\label{eq:objective_func_LBF}
\end{equation}
Comparing~\eqref{eq:objective_func} with~\eqref{eq:objective_func_LBF}, we have $R(\cdot)=||\Delta \mathbf{z}_i - \mathbf{w}_t \Phi(\mathbf{z}_i, I_i) ||_2^2$ and $\Omega(\cdot)=||\mathbf{w}_t||_2^2$. Consequently, we have the LBF algorithm embedded with our self-reinforcement.

The training of CFSS is to iteratively estimate a finer shape sub-region, $(\bar{\mathbf{x}}_{(l)}, P^R_{(l)})$, where $\bar{\mathbf{x}}_{(l)}$  is the center of the estimated sub-region and $P^R_{(l)}$ is the probability distribution depicting the sub-region around the center. We simply replace the regression stage in~\eref{eq:objective_func} with the iterative training of CFSS. At this moment, the regression parameter $\mathbf{w_t}$ indicates $(\bar{\mathbf{x}}_{(l)}, P^R_{(l)})$, and then we can apply the self-reinforced process for CFSS.

\section{Experimental Results and Analysis}
The experiments were performed on six widely used datasets include FRGC v2.0, LFPW, HELEN, AFW, iBUG and 300W. All faces are labeled 68 landmarks.
We compute the alignment error for testing images using the standard mean error normalized by the inter-pupil distance (NME). The value of error indicates the percentage of the inter-pupil distance, and we simply ignore the symbol `\%'.

Firstly, we verify the correlation between our discrepancy (no groundtruth label is available for its computation) and labeling error against the groundtruth. Then, we perform our self-reinforcement on two typical regressors and one recent deep model, resulting in examples of high quality at seven to twenty times, and finally compare our regression, whose training starts from a small number of labeled faces, with recent alignment algorithms.

\subsection{Correlation between discrepancy and error}

We analyze the effectiveness of discrepancy that evaluates the example goodness in our self-reinforcement. The discrepancy attemps to reflect the labeling error, i.e., how inaccurate a sample is labeled. Generally, samples exhibiting larger discrepancy have higher labeling error.

\begin{figure}[!htb]
\centering

\includegraphics[width=.8\linewidth]{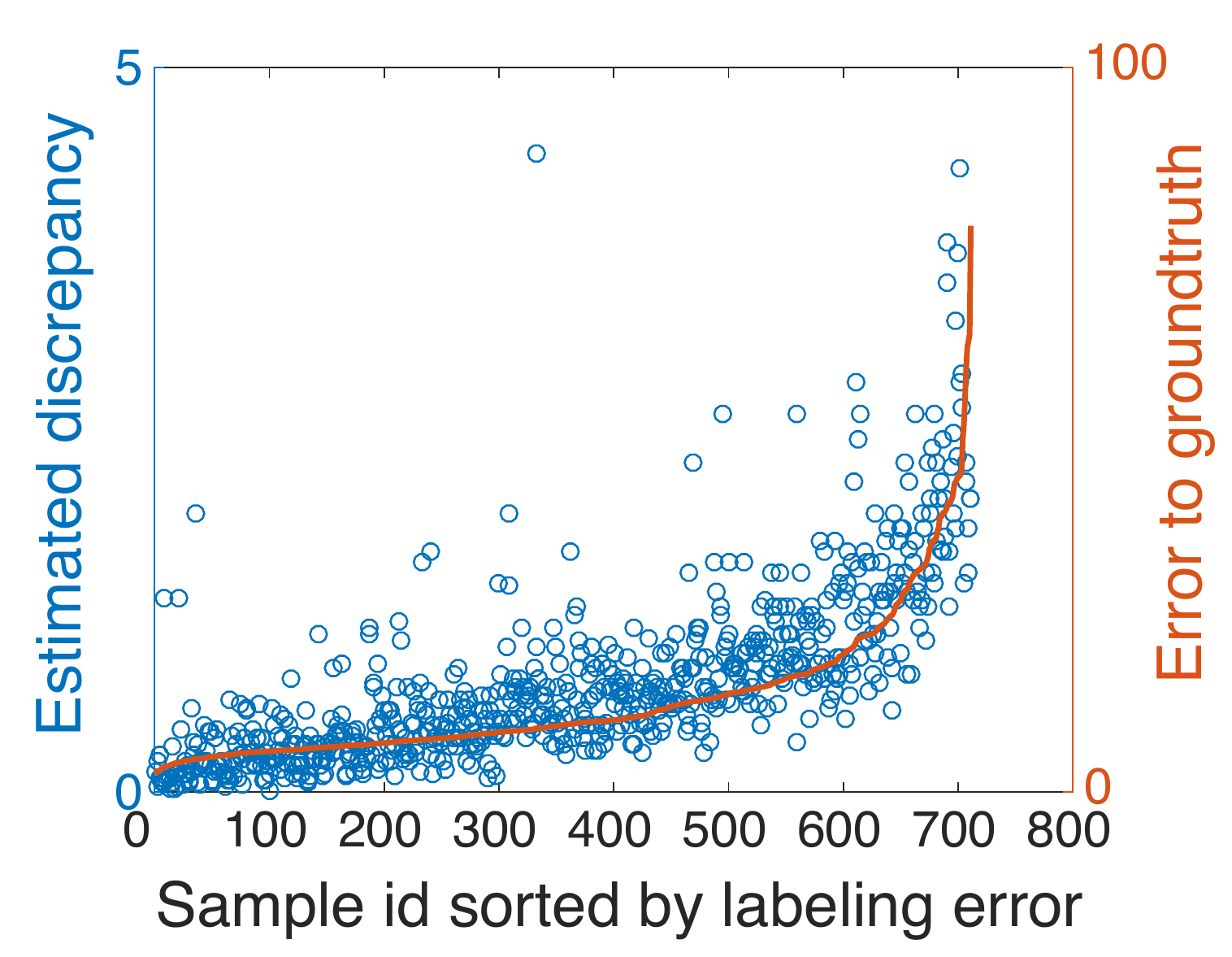}

\caption{The values of discrepancy and labeling error (the red line). One blue circle indicates the discrepancy value of one sample. }
\label{fig:error_v}
\end{figure}

To verify the correlation between the discrepancy and labeling error, we randomly chose 100 samples in LFPW, and trained an alignment regressor with these samples. Other 711 samples in LFPW were then labeled with the trained regressor. The labeling error and discrepancy of these predicted samples are plotted in~\fref{fig:error_v}.
The $x$ axis is sample IDs sorted by labeling error in an ascending order. The red line indicates the labeling error and one blue circle denotes the value of discrepancy for each sample. Figure~\ref{fig:error_v} demonstrates that there is a strong correlation between the discrepancy and labeling error. The values of the discrepancy for corresponding samples climb up with the increase of labeling error.
The red line fits the changes of the discrepancy very well. This fittingness verifies that the defined discrepancy reflects how accurate a label is. Therefore, every time we keep the samples having lower discrepancies, we have the most accurately labeled sample survived. These labels of low discrepancy introduce minimal error into training.

\subsection{Unlabeled example predicting and survival}

We firstly validate the self-reinforcement for typical regression, e.g., LBF and CFSS, on LFPW, and then our strategy for deep models highly data demanding on a larger mixed data set.
\subsubsection{Self-reinforcement on conventional regression}
\begin{figure}[!htb]
	\centering

	\includegraphics[width=.75\linewidth]{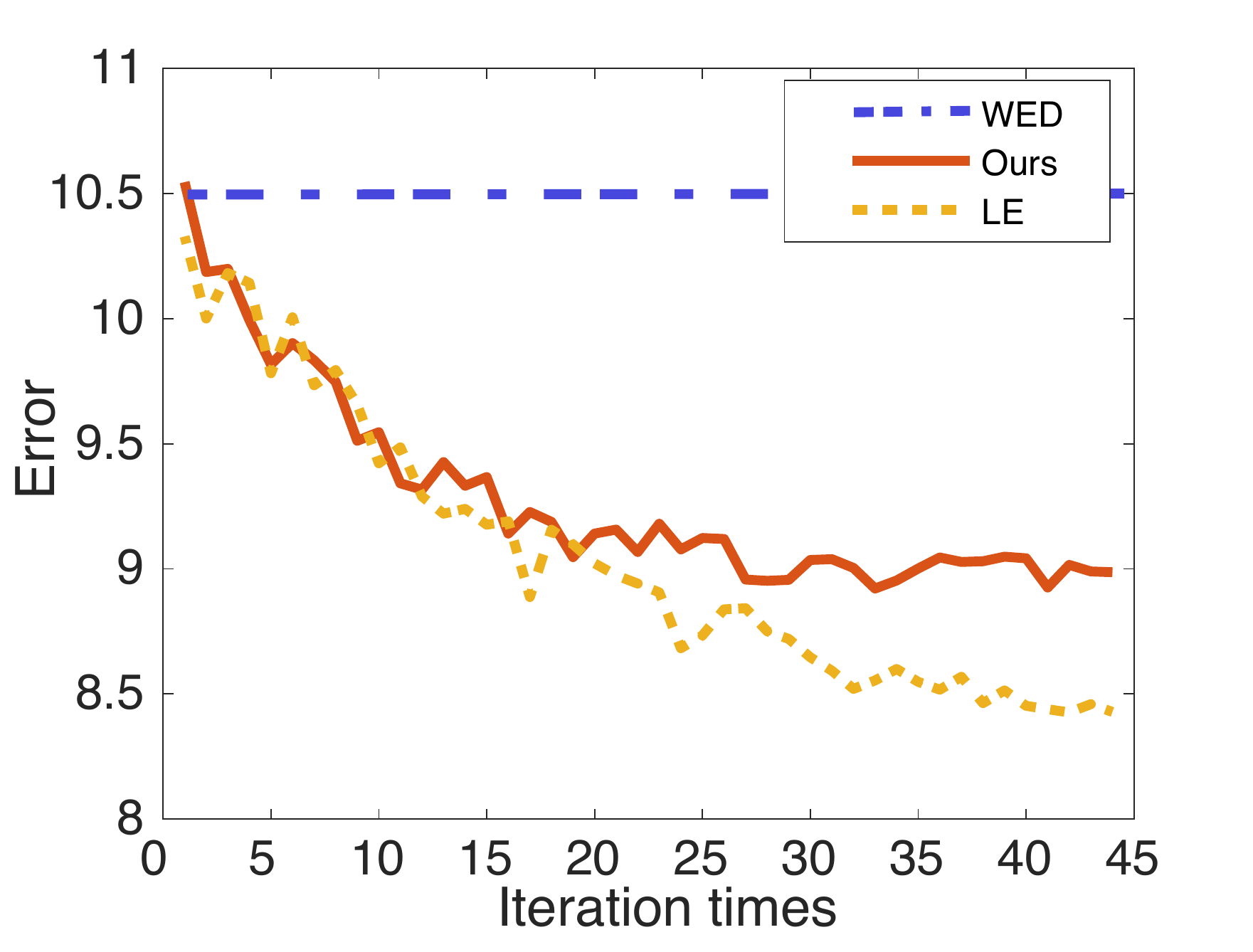}

	\caption{The test error of every iteration where the $y$ axis indicates the error, and $x$ denotes iteration steps. The blue dash dots are the errors of training without any extra unlabeled examples. The red solid and orange dots are those from ours and training with manually chosen examples.}
	\label{fig:error}
\end{figure}

LFPW contains more than one thousand images showing great variations especially on pose changes. Previous studies show that LBF and CFSS perform well on this set as long as hundreds of accurately labeled faces are available. We validate how close the self-reinforced versions of LBF and/or CFSS with unlabeled examples work to the original algorithms with labeled ones.

Firstly, we validate how the minimization of our objective~\eref{eq:objective_func} continuously predicts and preserves those examples of low discrepancy. Manually including examples of the lowest prediction error against the groundtruth (available in LFPW) gives the upper bound of the example survival. We started from 100 labeled examples, and implemented the self-reinforced version of LBF (SR-LBF) to automatically include 711 extra samples (regarded as unlabeled). The comparisons between manual inclusion of the lowest labeled error (LE) and our SR-LBF are plotted in~\fref{fig:error} showing the mean alignment error in every iteration.
The testing error of SR-LBF on 224 images, shown as the red solid line, decreases from
10.5 to 8.98, 14\% lower than training without any extra unlabeled data (WED). The orange dots indicate the alignment errors of the regression with manually chosen samples having the lowest labeling error against the groundtruth. There is almost no difference between ours and LE in the beginning of the iteration process. The gap increases as more self-reinforced samples, automatically labeld and survived, are included, but reaches as low as 0.5 when the process converges. Our self-reinforcement is not necessarily able to generate and include the `groundtruth' labels (not exist in practice), but definitely to improve the behavior of the regression toward the optimal.

\begin{figure}[h]
	\begin{center}
		\includegraphics[width=.9\linewidth]{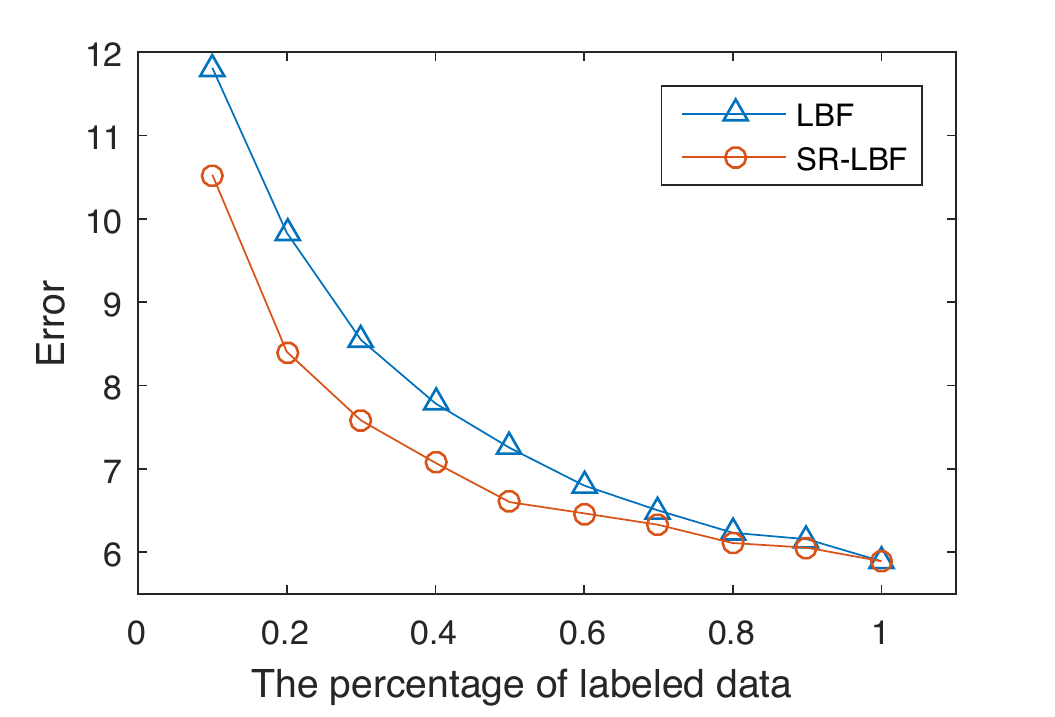}%
	\end{center}
	\caption{The mean alignment error under different ratios of manually groundtruth labels. The orange circles give the errors of SR-LBF with the mixture of groundtruth and automatically labeled examples by self-reinforcement. The blue triangles are those from LBF trained by various portions of groundtruth labels indicated by the $x$ axis.}
	\label{fig:different_ratio}
\end{figure}

Secondly, we demonstrate the effectiveness of self-reinforcement by comparing SR-LBF with LBF when including different ratios of groundtruth labels for training. Besides those groundtruth labels, SR-LBF can include the rest of LFPW training images without their labels. Figure~\ref{fig:different_ratio} illustrates the mean errors for SR-LBF and LBF on 224 testing LFPW images. As the increase of the percentage of groundtruth labels, both LBF and SR-LBF give lower errors because the quantity of training examples with high quality labels is expanding. The errors of SR-LBF are always lower than LBF, and the gaps are evident especially when only small fractions (less than 50\%) of groundtruth labels are available. When all groundtruth labels are given, our regression degrades to LBF. This plot validates that the self-reinforcement is able to expand the quantity of training examples while maintaining the quality.
\begin{figure}[!htb]
\centering
\includegraphics[width=.8\linewidth]{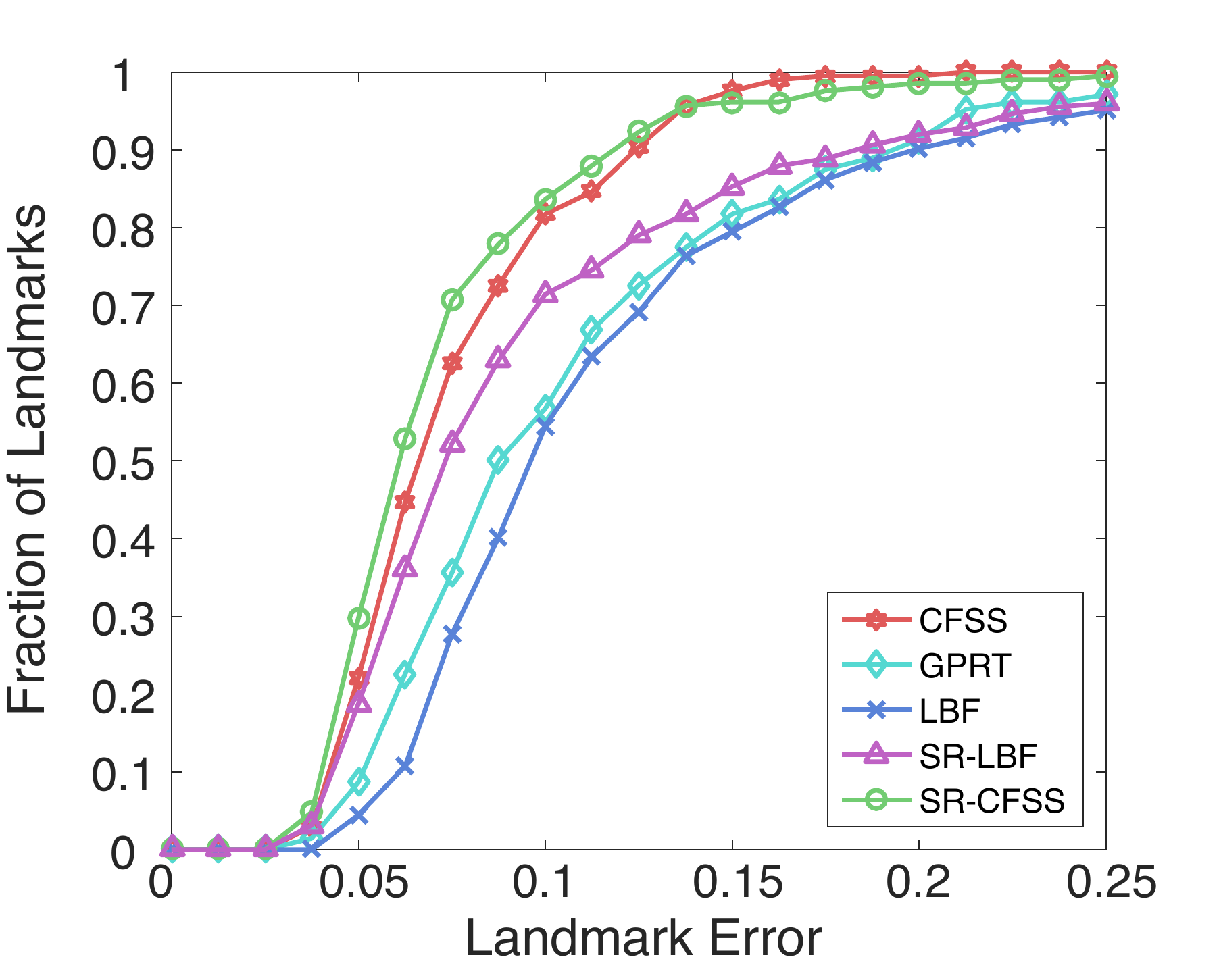}
\caption{Cumulative errors distributions on testing images of LFPW. The $x$-axis is the normalized mean error (NME), and the $y$-axis indicates the percentage of images on which NMEs are lower than the $x$ value.}
\label{fig:error_1}
\end{figure}

\begin{figure}[!htb]
\centering
\includegraphics[width=.8\linewidth]{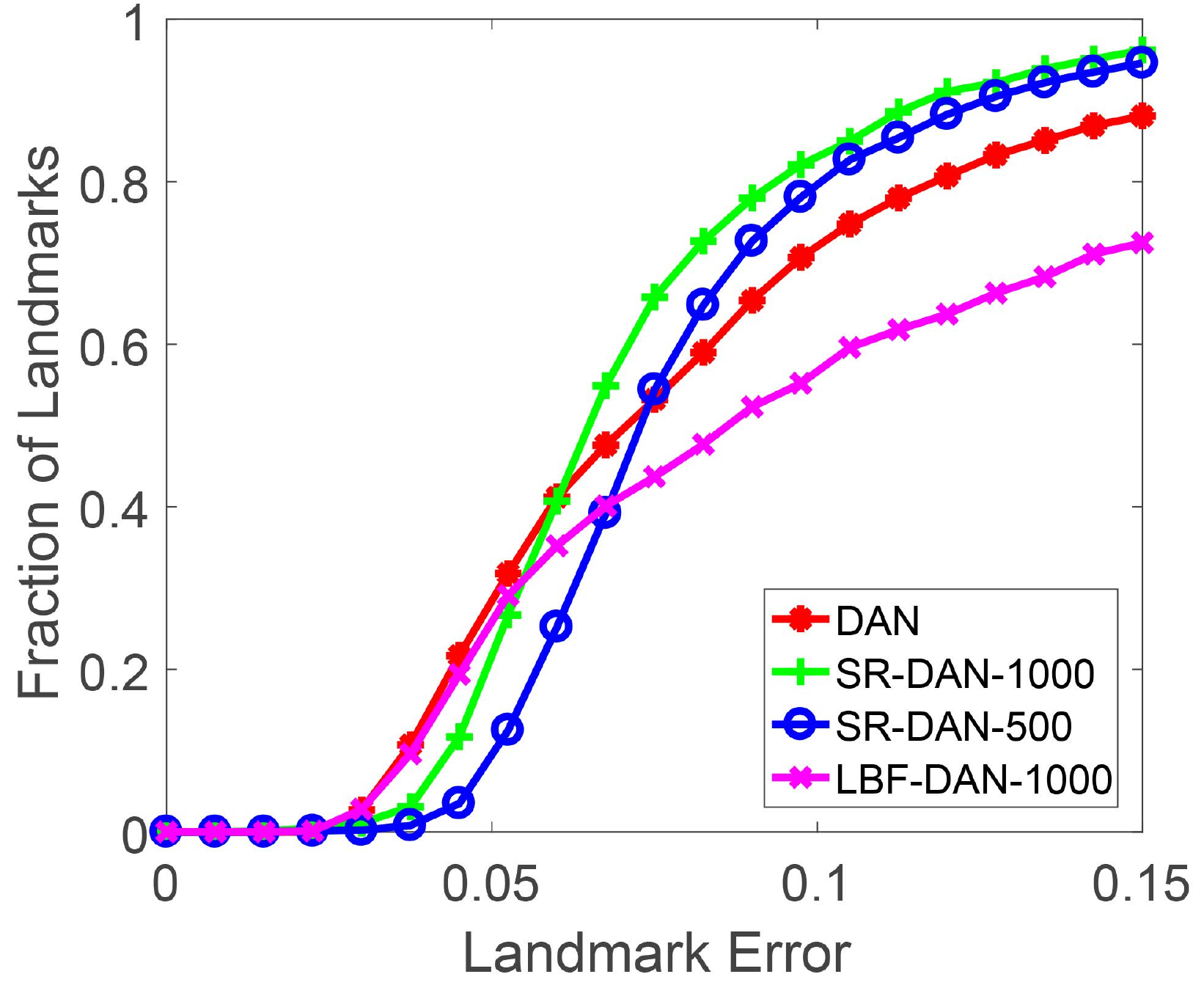}
\caption{Cumulative errors distributions on Large dataset. The $x$-axis is the normalized mean error (NME), and the $y$-axis indicates the percentage of images on which NMEs are lower than the $x$ value.}
\label{fig:error_2}
\end{figure}

Thirdly, we compare the self-reinforced versions of CFSS~\cite{zhu2015face} and LBF~\cite{ren2014face} with the original algorithms as well as GPRT~\cite{lee2015face}. Figure~\ref{fig:error_1} illustrates the cumulative error distribution plots on 224 testing images of LFPW. All methods were trained with only 100 annotated images, but our self-reinforcement included 711 extra unlabeled samples. SR-CFSS has better performance than CFSS, and SR-LBF better than LBF. Both perform superior than GPRT, and SR-CFSS is the best of these five algorithms. The proposed self-reinforcement is capable of automatically labeling examples and preserving good ones. Faces annotated with alignment results are shown in~\fref{fig:results}\footnote{More images are available in the supplementary materials}. The SR versions performs much better on noses and mouthes presenting large variations that cannot be covered by a small number of training examples in the original regression algorithms.
\begin{figure}[h]
\centering
\subfigure[LFPW]{
\includegraphics[width=.45\linewidth]{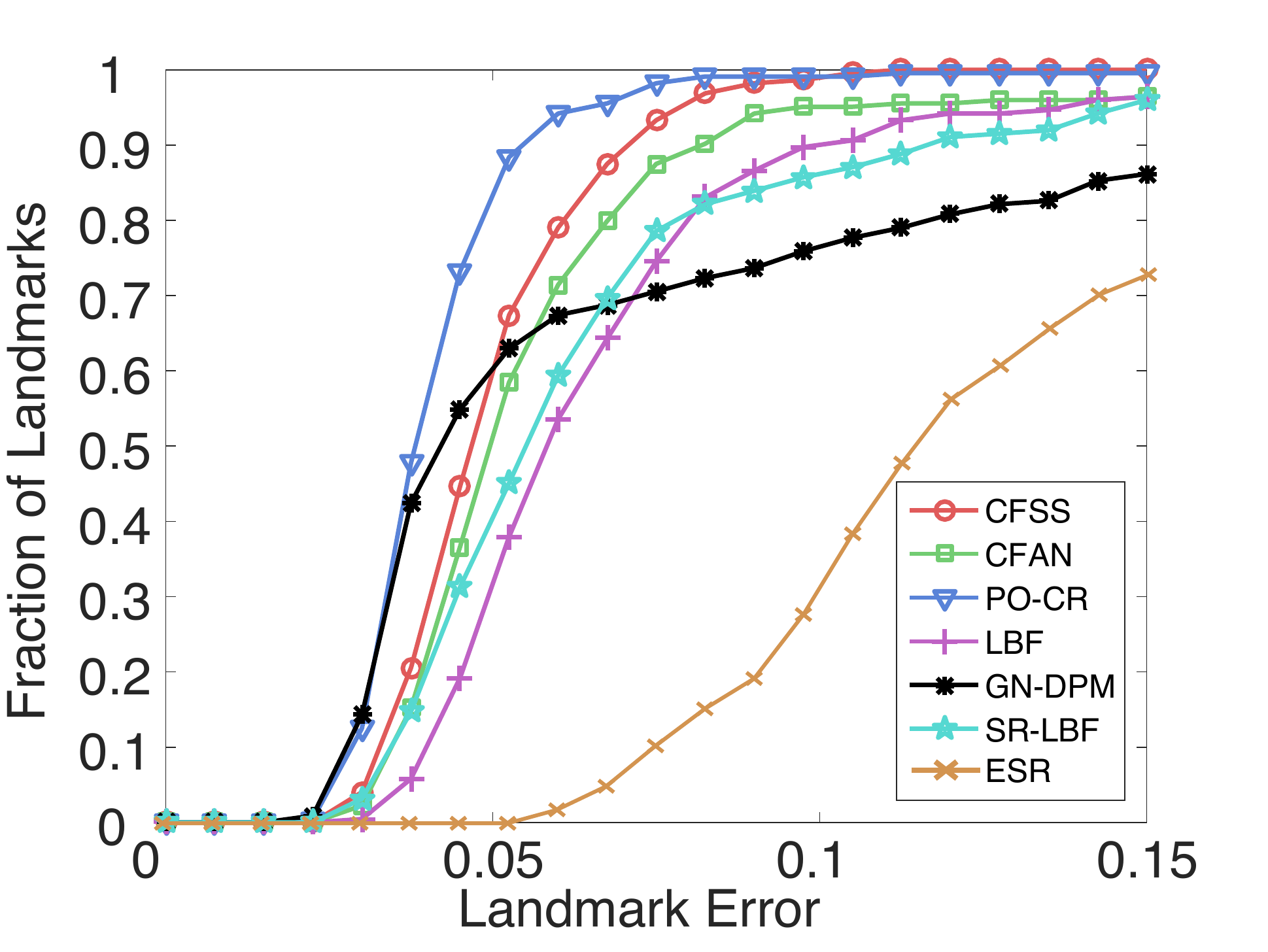}
}
\subfigure[Helen]{
\includegraphics[width=.434\linewidth]{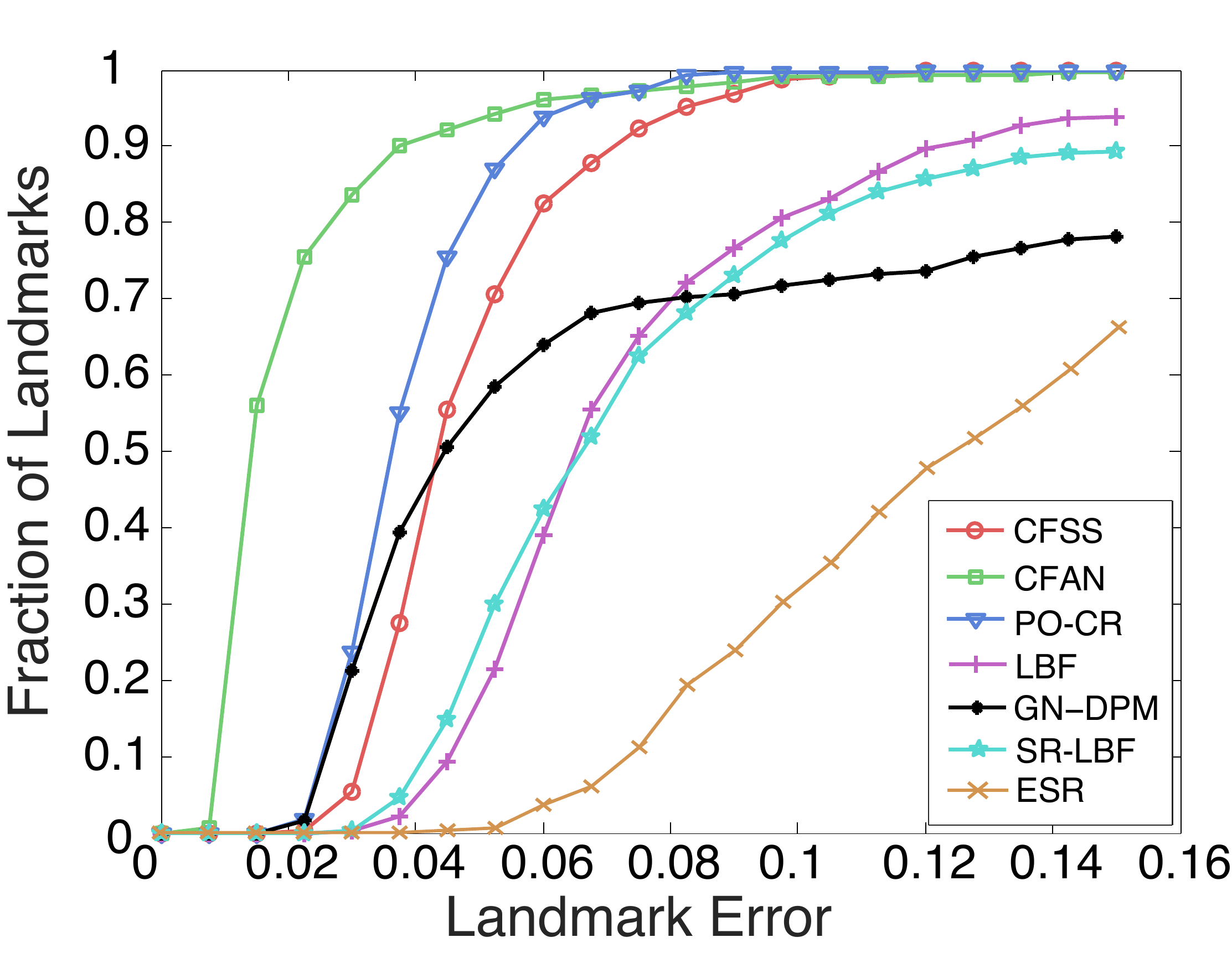}
}

\caption{Cumulative errors distributions tested on LFPW and Helen.}
\label{fig:300w}

\end{figure}

\subsubsection{Self-reinforcement on deep networks}
To test the capability of our self-reinforced strategy on a large amount of unlabeled facial images, we construct a large dataset which contains 8,151 images and is made up of 6 facial datasets include FRGC v2.0, LFPW, HELEN, AFW, IBUG and 300W. We compare the performance between the DAN\cite{kowalski2017deep} trained only by labeled examples, labeled examples with extra examples obtained by our self-reinforced strategy and labeled examples with extra examples obtained by LBF. The number of labeled examples is 100. Our self-reinforced framework use LBF as alignment algorithm and obtains over 3,000 labeled facial images (some bad samples are not chosen), then we choose 400 and 900 of them as extra examples for DAN. We also directly run LBF~\cite{ren2014face} which is trained by 100  samples on the large dataset, then perform randomly selection on the result of LBF to obtain 900 extra examples for DAN. 1,000 images from the large dataset are used for testing. Figure~\ref{fig:error_2} illustrates the cumulative error distribution of these methods. As a deep learning method, DAN needs large amount of training data. The result shows that, when there are only 100 labeled training data provided, our method can enhance the performance of DBN by provide them another 400 training data. The performance can be improved when the number of extra data is added from 400 to 900.  The comparisons between the regressor trained by labeled examples with extra examples obtained by our self-reinforced strategy and labeled examples with extra examples obtained by LBF prove that: selecting extra samples indiscriminately cannot only improve the performance but also result in poor accuracy.


\subsection{Quantitative comparisons with the state-of-the-art}

We conducted comparisons with six face alignment algorithms on 300-W. These six face alignment regressors are pre-trained by a \emph{huge} number of labeled images. CFAN and CFSS were trained on a combination of Helen (2000), LFPW (811) and AFW (337). The total number of these training samples is 3148. PO-CR and GN-DBM~\cite{tzimiropoulos2014gauss} were trained on the training set consisting of LFPW and Helen. ESR~\cite{cao2014face} were trained on Helen. The total number of training samples is 2811. GPRT and LBF were trained on the training set of LFPW having 811 labeled images. In contrast, our self-reinforced LBF (SR-LBF) starts from only a half of LFPW, i.e. 400 training labels, and the other half are included by our self-reinforced strategy. The cumulative error distributions of the compared methods and ours are shown in Figure~\ref{fig:300w}.

The comparisons show that our regression does not necessarily give a better performance than the others. Instead, we are able to achieve comparable performance on common subsets of 300-W with an extremely small training set of labels. The number of our training labels is one half of GPRT, 25\% of ESR, 14\% of PO-CR and GN-DBM, and only 12\% of CFAN and CFSS. Especially, our regression performs close to LBF with half of labels. Again, our self-reinforcement is open to any cascaded regression, and has the potential to improve the respective ability by automatically predicting and preserving high quality labels.

\section{Conclusion}
We propose a self-reinforced cascaded regression that fuses the discovering and upgrading training examples of low discrepancy into cascaded regression for face alignment. The framework is derived from indirect consistency with local appearance and global geometry.
Finally, we validate the effectiveness of our regression. We are not intending to devise a competitive alignment algorithm trained with huge collected labels, but instead a self-reinforced strategy that automatically expands good training examples from a small subset, thus being complementary and more general to existing cascaded regression.
\begin{figure}[H]
	\begin{center}
		\includegraphics[width=0.9\linewidth]{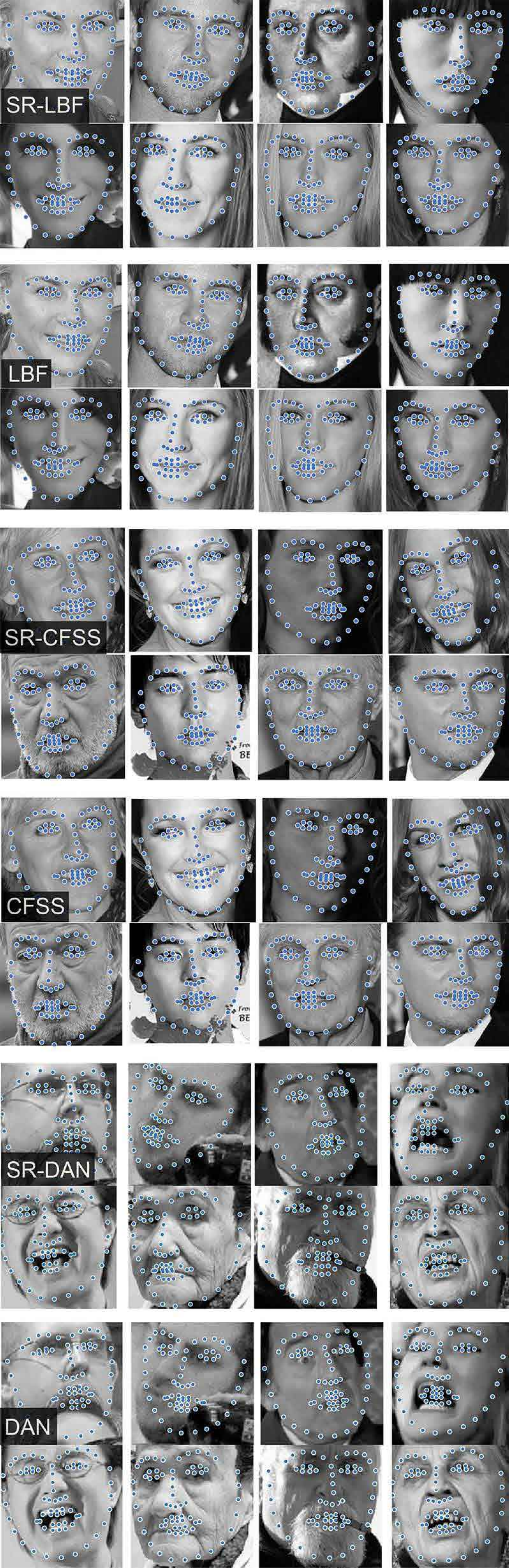}%
	\end{center}
	\caption{Example images from LBF, SR-LBF, CFSS, LR-CFSS on LFPW; Example images from DAN and LR-DAN on large dataset}
	\label{fig:results}
\end{figure}

\section{Acknowledgments}

This work is partially supported by the National Natural Science Foundation of China (Nos. 61572096, 61432003, 61733002, 61672125, and 61632019), and the Hong Kong Scholar Program (No. XJ2015008). Dr. Liu is also a visiting researcher with Shenzhen Key Laboratory of Media Security, Shenzhen University, Shenzhen 518060.

\bibliographystyle{aaai}
\bibliography{Fan}
\end{document}